\documentclass{article}
\usepackage{spconf,amsmath,graphicx}

\usepackage{enumitem}
\setlist{nosep, leftmargin=14pt}

\usepackage{booktabs}
\usepackage{multirow}

\usepackage{soul}
\usepackage{url}

\usepackage{caption}
\usepackage{subfig}
\usepackage{graphicx}
\usepackage{svg}

\usepackage[normalem]{ulem} 
\useunder{\uline}{\ul}{}

\title{Improving segmentation of retinal arteries and veins using cardiac signal in Doppler Holograms}
\name{
  \begin{minipage}{\linewidth}
    \centering
  Marius Dubosc$^{\star}$ \qquad Yann Fischer$^{\dagger}$ \qquad Zacharie Auray$^{\dagger}$ \\
Nicolas Boutry$^{\star}$ \qquad Edwin Carlinet$^{\star}$ \qquad Michael Atlan$^{\dagger}$ \qquad Thierry
Géraud$^{\star}$
\end{minipage}
}

\address{$^{\star}$ EPITA Research Laboratory, \textit{Le Kremlin-Bicêtre, France} \\
    $^{\dagger}$ Langevin Institute, \textit{Paris, France}
}
\begin{document}
%\ninept
%

\maketitle

\begin{abstract}

Doppler holography is an emerging retinal imaging technique that captures the dynamic behavior of blood flow with high temporal resolution, enabling quantitative assessment of retinal hemodynamics. This requires accurate segmentation of retinal arteries and veins, but traditional segmentation methods focus solely on spatial information and overlook the temporal richness of holographic data. In this work, we propose a simple yet effective approach for artery–vein segmentation in temporal Doppler holograms using standard segmentation architectures. By incorporating features derived from a dedicated pulse analysis pipeline, our method allows conventional U-Nets to exploit temporal dynamics and achieve performance comparable to more complex attention- or iteration-based models. These findings demonstrate that time-resolved preprocessing can unlock the full potential of deep learning for Doppler holography, opening new perspectives for quantitative exploration of retinal hemodynamics.
The dataset is publicly available at
https://huggingface.co/datasets/DigitalHolography/

% This study introduces a novel approach for segmenting retinal arteries and veins in temporal Doppler holograms using a standard U-Net architecture. The method leverages a dedicated pulse signal analysis pipeline to extract informative temporal features, enabling conventional U-Nets to implicitly exploit temporal dynamics and compete with more complex iterative and attention-based models, surpassing state of the art semantic segmentation of retinal vessels on traditional datasets. It further enables the quantitative estimation of hemodynamic biomarkers, such as blood flow, from real-time Doppler holography acquired at 37,000 frames per second.
% Our findings demonstrate that U-Nets can effectively capture temporal information when properly preprocessed, paving the way for quantitative analyzes based on Doppler holography. This work highlights the potential of Doppler holography and the richness of its temporal and quantitative information, which offers new opportunities for deep learning–based exploration of retinal hemodynamics. 
\end{abstract}

\section{Introduction}
\label{sec:intro}

The retina provides a unique window into microvascular health, with blood flow and vessel morphology reflecting diseases such as diabetic retinopathy, glaucoma, hypertension, and Alzheimer’s. Recent advances in retinal imaging have improved visualization of the vascular network, yet most modalities remain limited to static or qualitative flow representations. Fluorescein and OCT angiography map vessel topology with high resolution, but rely on surrogate markers and cannot capture temporal dynamics throughout the cardiac cycle. Laser Doppler flowmetry, laser speckle contrast imaging, and ultrasound Doppler measure local flow but lack sufficient spatial resolution to resolve retinal layers or specific vessels~\cite{fischer_retinal_2024}. Doppler holography, by contrast, enables simultaneous imaging of vessel morphology and hemodynamics at high frame rates~\cite{dubosc_holovibes_2025}, directly quantifying Doppler shifts induced by moving red blood cells and providing time-resolved measurements of flow velocity and direction over a wide field~\cite{fischer_retinal_2024}.

Extracting quantitative hemodynamic information requires semantic segmentation of retinal arteries and veins, a challenging task due to overlapping intensity distributions, variable vessel calibers and image noise. Traditional approaches rely on hand-crafted features such as vessel color, width, or proximity to the optic disc, while modern methods increasingly use deep learning architectures, particularly U-Net variants. Many are fine-tuned for fundus images, which exhibit a wide range of vessel calibers and resolutions. To handle these multiscale challenges, some methods adopt iterative strategies, specialized convolution layers, spatial attention, or graph-based networks.

Most of these spatially tuned strategies, however, lose effectiveness with Doppler holography. Power Doppler images~\cite{puyo_spatio_2020} have lower vessel contrast and resolution, fewer small vessels, and overlapping retinal and choroidal vasculature. In this context, temporal information becomes the most informative dimension. By analyzing the pulse signal over time, arteries and veins can be reliably distinguished. Using this approach, we train several models and achieve highly satisfactory results, with simple U-Net-like architectures performing on par with more complex iterative, attention-based, or Transformer-based networks.

\begin{figure}[t]
\centering
\subfloat[M0 image]{\includegraphics[width=2.7cm]{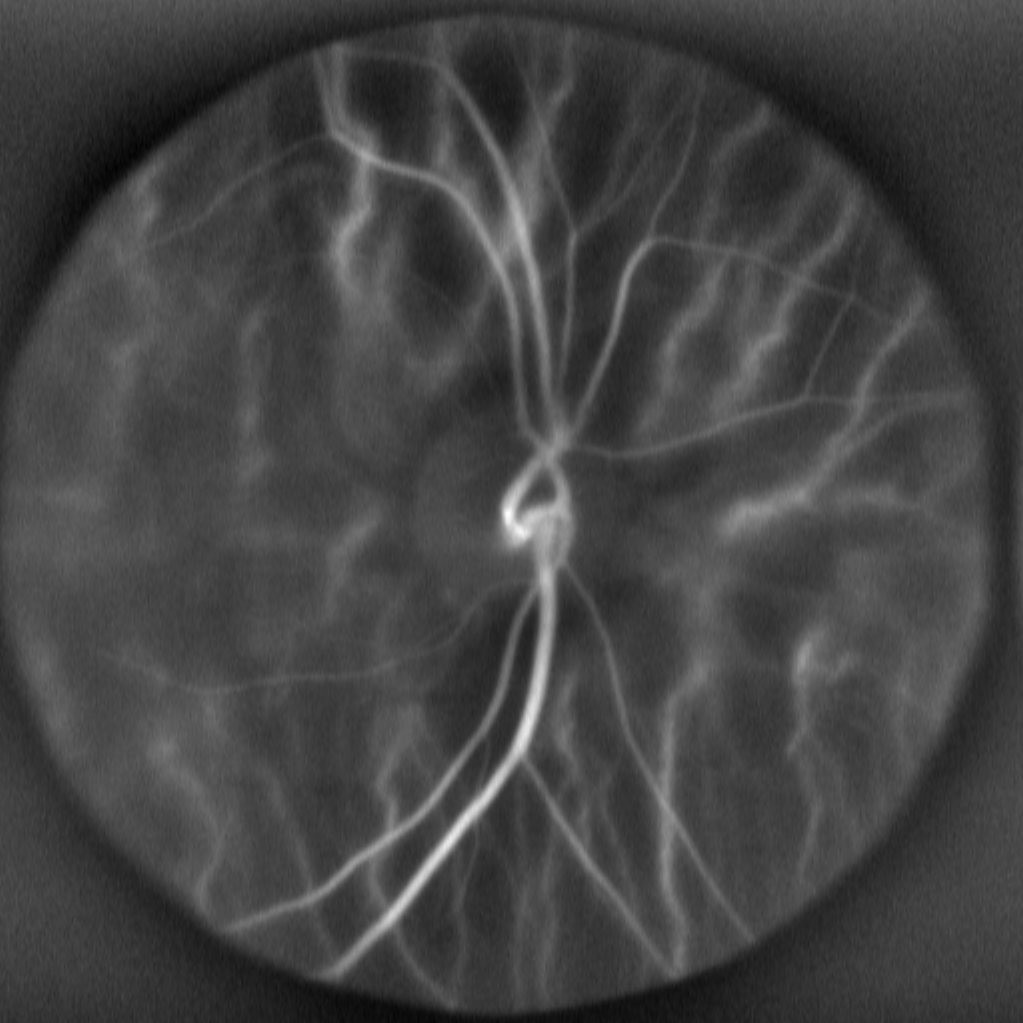}}\hfil 
\subfloat[Correlation map]{\includegraphics[width=2.7cm]{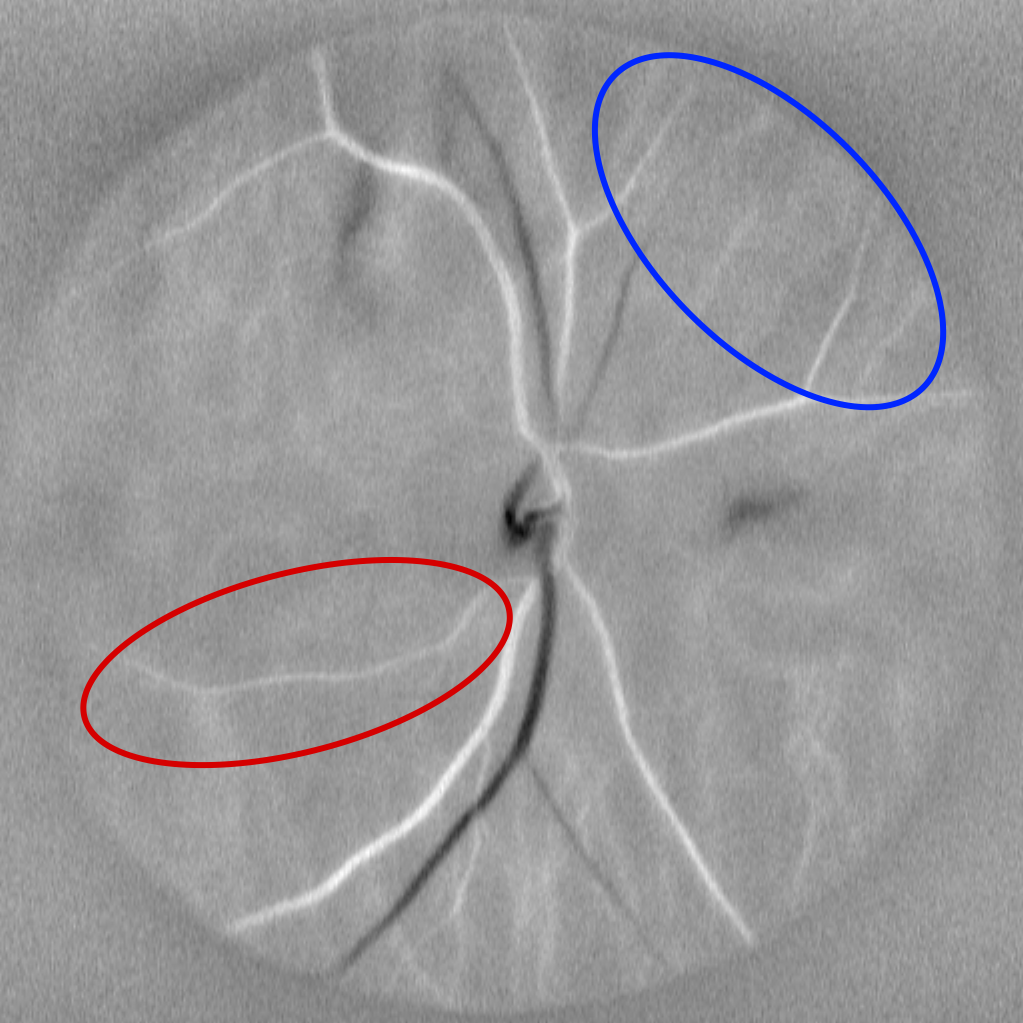}}\hfil   
\subfloat[Diasys image]{\includegraphics[width=2.7cm]{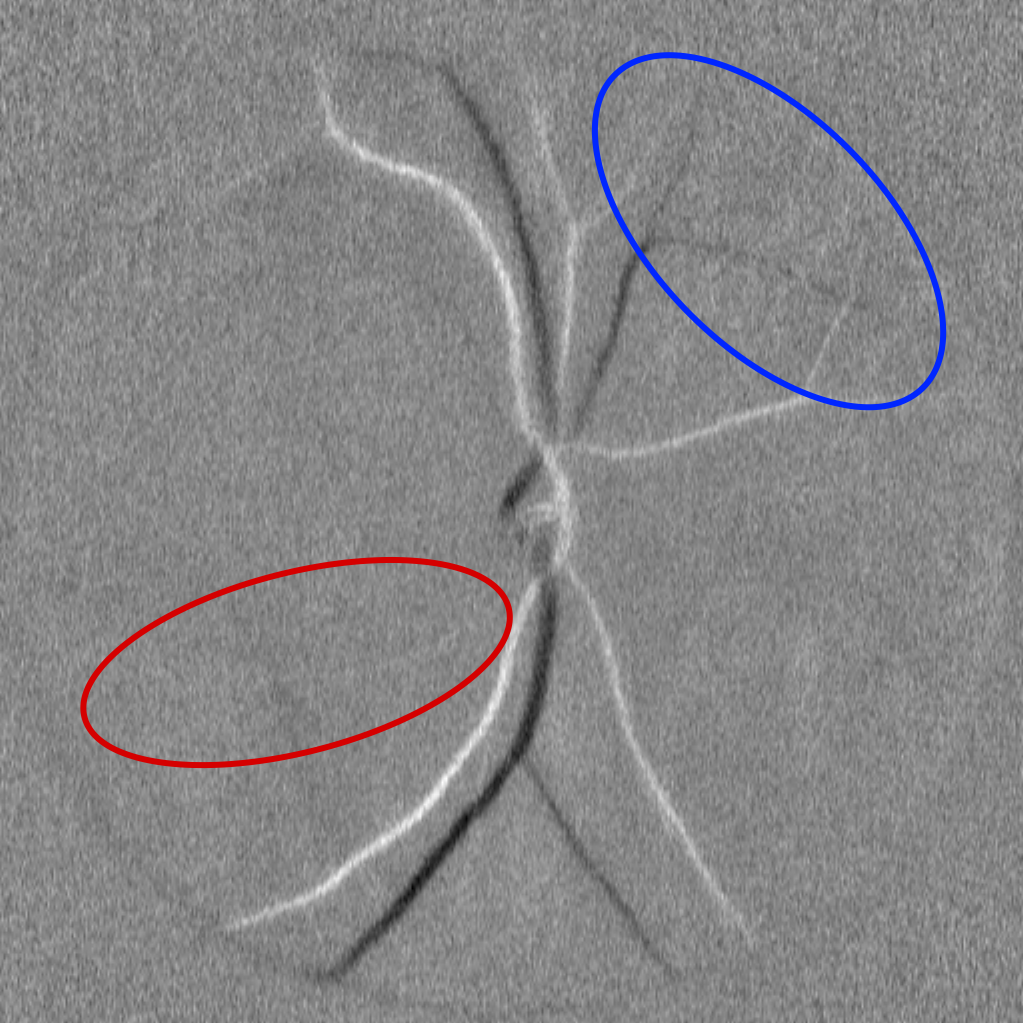}}\hfil
\caption{Different processing of the M0 video. Its average M0 image (a) looses temporal fluctuations, but offers better visualization of the vasculature, including choroidal vessels, not connected to the centered optic disc.
Correlation map (b) and Diasys image (c) both convey retinal hemodynamic information, but they respectively better reveal arteries and veins, as shown by red (artery) and blue (vein) selections.}\label{inputs}
\end{figure}

\section{Doppler holography}

Doppler Holography exploits Doppler shifts induced by moving red blood cells to reveal the vascular network and quantify blood flow velocity and direction.

Using a Mach-Zehnder inline interferometer, a diffuse laser beam illuminates the retina. Its backscattered light interacts with a reference beam, creating interferogram patterns acquired by a high-speed camera. The current device setup is depicted in \cite{fischer_retinal_2024}, while \cite{bratasz_diffuse_2022} explains the experimental protocol.

The raw 512x320-pixel interferograms are acquired at 37.000 fps using the dedicated open-source acquisition and real-time rendering software \textit{Holovibes} \cite{dubosc_holovibes_2025}. Offline rendering of the raw interferograms is then performed using the open-source Matlab program \textit{Holodoppler} \cite{Holodoppler}. The interferograms are reconstructed at the retina plane via Fresnel propagation, then eye-motion artifacts are removed using singular value decomposition filtering \cite{puyo_spatio_2020}. High-pass filtering of the temporal frequency spectrum obtained from the short-time Fourier transform (512-frame windows) suppresses low-frequency components from static tissue and eye motion, retaining the frequency shifts associated with moving red blood cells \cite{puyo_vivo_2018}.

The Doppler Power Spectrum Density (DPSD) is obtained by computing the squared magnitude of the filtered Fourier spectrum for each time window. The moment of order zero of this spectrum (denoted \textit{M0}), obtained by integrating the DPSD over all frames, enables the visualization of blood flow dynamics across the cardiac cycle. It is the data used throughout our pipeline. The accumulation of each frame produces Power Doppler images - \textit{M0 images} (see figure~\ref{inputs}) - providing a high-resolution map of the retinal vasculature, but losing its temporal fluctuations. 

Using a light scattering model, blood flow velocity in retinal arteries and veins is calculated by comparing local DPSD broadening in vessels with neighboring tissue \cite{fischer_retinal_2024}, which requires semantic segmentation of arteries and veins.

\begin{figure*}[t]
    \centering
    \includegraphics[width=\linewidth]{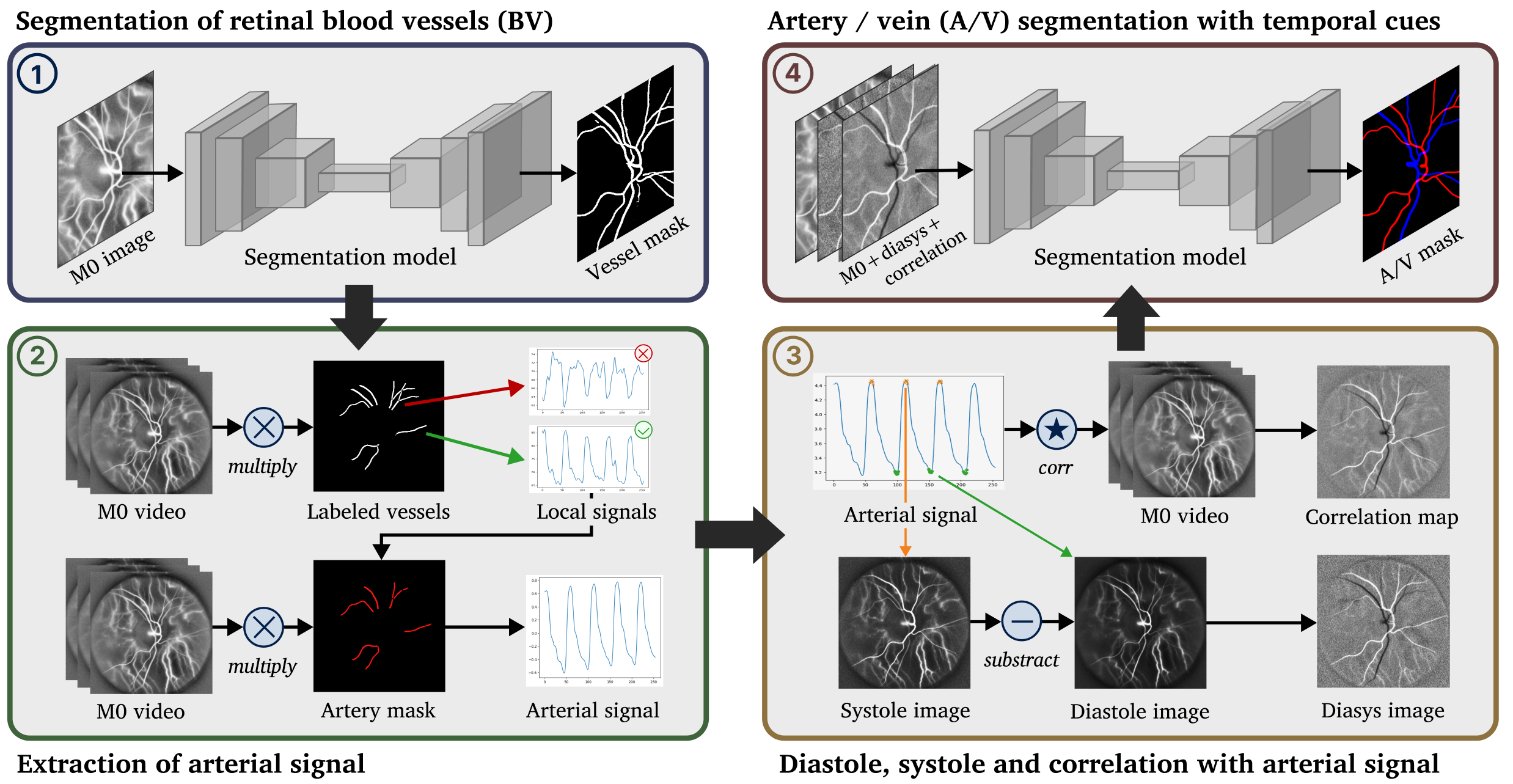}
    \caption{Sequential segmentation pipeline.
    1. Segmentation of retinal blood vessels, using a classic retinal segmentation model
    2. By analyzing the local signal of each labeled vessel from the binary vessel mask, arteries with the most dominant signal are identified.
    3. Using the arterial signal, temporal cues are extracted: zero lag cross-correlation with the Power Doppler video gives a correlation map with the arterial signal for each pixel, and the peaks and valleys are used to extract the diastolic and systolic frames.
    4. The temporal cues are concatenated with the Power doppler (M0) image, and given as input to a classic artery / vein segmentation model.}
    \label{fig:segmentation}
\end{figure*}

% \section{Segmentation strategies}

% \begin{figure*}[!t]
% \centering
% \subfloat[Power Doppler image (M0)]{\includegraphics[width=5cm]{figures/250507_COY_L_2_HD_1_M0.png}}\hfil
% \subfloat[B]{\includegraphics[width=5cm]{figures/mask_vesselness.png}}\hfil 
% \subfloat[C]{\includegraphics[width=5cm]{figures/250507_COY_L_2_HD_1_EF_3_artery_16_PreMask.png}} 

% \subfloat[D]{\includegraphics[width=5cm]{figures/correlation.png}}\hfil   
% \subfloat[E]{\includegraphics[width=5cm]{figures/diasys.png}}\hfil
% \subfloat[F]{\includegraphics[width=5cm]{figures/250507_COY_L_2_HD_1_M0.png}}
% \caption{Many figures}\label{figure}
% \end{figure*}

% \begin{figure*}[!t]
%      \centering
%      \begin{subfigure}[b]{0.3\textwidth}
%          \centering
%          \includegraphics[width=\textwidth]{figures/250507_COY_L_2_HD_1_M0.png}
%          \caption{$y=x$}
%          \label{fig:y equals x}
%      \end{subfigure}
%      \hfill
%      \begin{subfigure}[b]{0.3\textwidth}
%          \centering
%          \includegraphics[width=\textwidth]{figures/mask_vesselness.png}
%          \caption{$y=3\sin x$}
%          \label{fig:three sin x}
%      \end{subfigure}
%      \hfill
%      \begin{subfigure}[b]{0.3\textwidth}
%          \centering
%          \includegraphics[width=\textwidth]{figures/250507_COY_L_2_HD_1_EF_3_artery_16_PreMask.png}
%          \caption{$y=5/x$}
%          \label{fig:five over x}
%      \end{subfigure}
%         \caption{Three simple graphs}
%         \label{fig:three graphs}
% \end{figure*}

\section{Segmentation of retinal arteries and veins}

\subsection{Dataset and models}
The segmentation pipeline takes as input the rendered power Doppler videos and their averaged image, denoted M0. We used a local dataset of 145 samples from 47 patients, with handcrafted artery/vein masks, to train and evaluate various traditional and state-of-the-art segmentation models. The annotations were produced by a non-expert PhD student (first author), relying on the temporal cues described in subsection \ref{sequential}. The data were split into 121 training and 24 test samples, with no patient overlap between splits. Due to the recent nature of the dataset, some patients have multiple measures, which may introduce bias; however, experiments showed that using all available data yielded the best performance.

The models used in this study mostly represent successive evolutions of the U-Net architecture for semantic segmentation, differing mainly in how they enhance basic convolutional units, integrate multi-scale context, refine predictions iteratively, or capture global dependencies.

The U-Net \cite{ronneberger2015u} serves as the foundation, combining an encoder–decoder structure with skip connections for end-to-end segmentation. U-Net++ \cite{zhou2018unet++} enriches feature fusion with dense skip connections. MSU-Net \cite{su_msunet_2021} extends context modeling through multi-scale convolutions, and CE-Net \cite{gu_cenet_2019} captures high-level semantics from multiple scales via dilated and pooling modules.

Sequential architectures refine predictions iteratively: IterNet \cite{li_iternet_2020}, W-Net \cite{galdran_minimalistic_2022} and RRWNet \cite{morano_rrwnet_2024} use cascaded or recursive subnetworks to correct vessel discontinuities and enforce topological consistency. 

Transformer-based models such as Swin-Unet \cite{cao_swinunet_2022} and hybrid designs like \st{W}Net \cite{zhou_nnwnet_2025} leverage self-attention to encode long-range dependencies, unifying global context with local detail preservation. Finally, CSTA-NET \cite{shabani2025coupled} combines Swin Transformers with the sequential architecture approach, using convolutions and fusion blocks to combine global and local feature maps.

All models were trained with the AdamW optimizer (\(\beta_1\) = 0.9, \(\beta_2\) = 0.99, weight decay = \(10^{-2}\)). Fine-tuning was performed for 300 epochs using the 1Cycle learning rate policy~\cite{smith2018disciplined} with a maximum learning rate of \(10^{-2}\). An early stopping criterion was applied with a patience of 30 epochs and a minimum improvement threshold of \(10^{-6}\) to prevent overfitting.
The cross-entropy loss was used for all models except RRWNet, which was trained with its custom RRLoss~\cite{morano_rrwnet_2024}.

\begin{figure}[t]
\centering
% Column 1
\subfloat[Ground truth]{
\begin{minipage}[c]{2.7cm}
    \frame{\includegraphics[width=\linewidth]{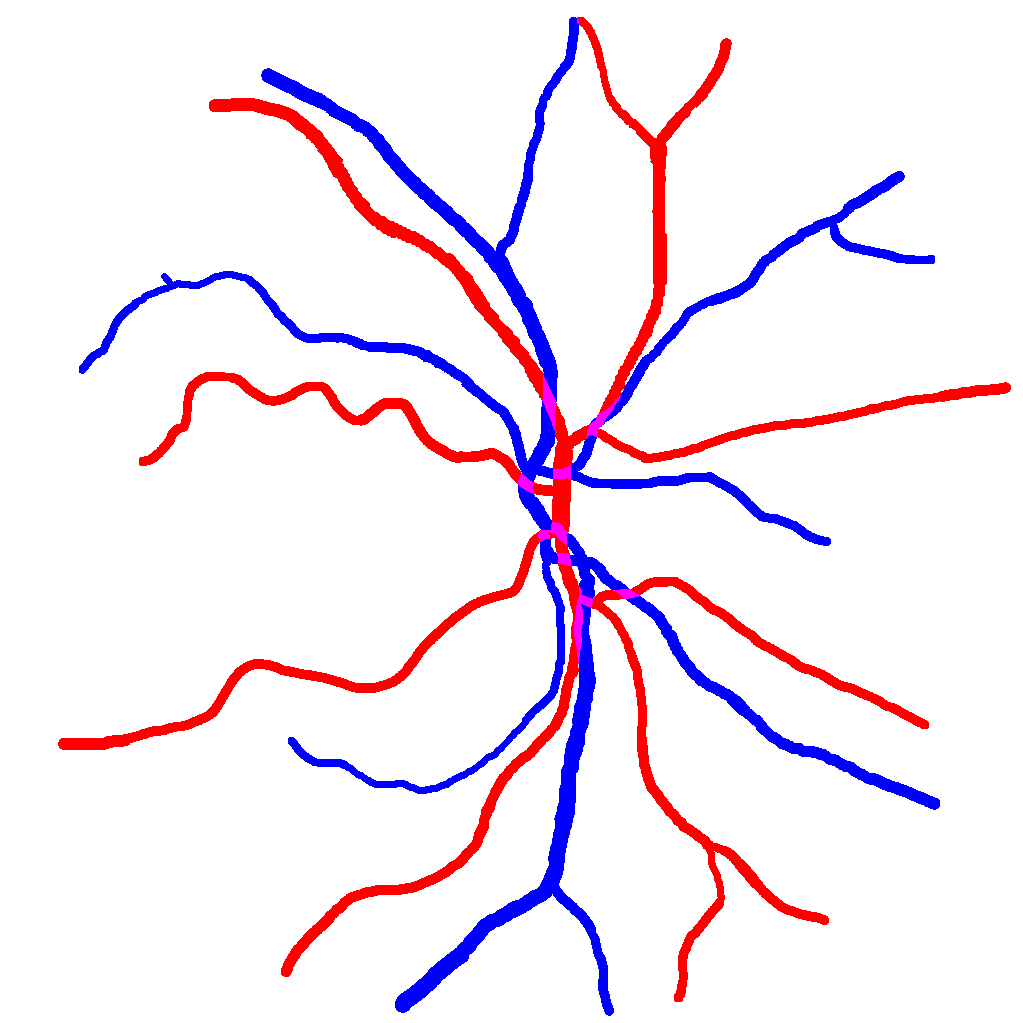}}\\[0.3em]
    \frame{\includegraphics[width=\linewidth]{figures/target_mask_white.png}}
\end{minipage}
}\hfil
% Column 2
\subfloat[M0 input]{
\begin{minipage}[c]{2.7cm}
    \frame{\includegraphics[width=\linewidth]{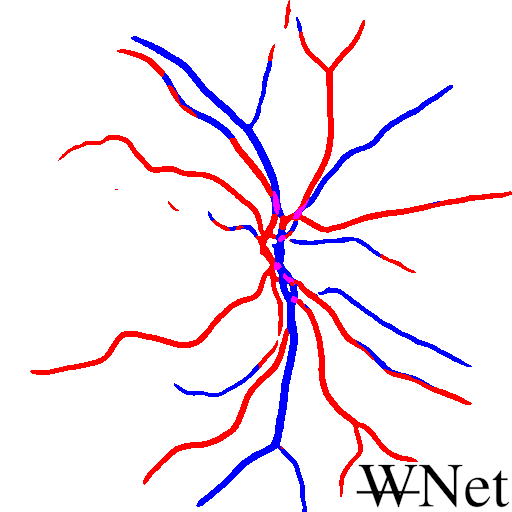}}\\[0.3em]
    \frame{\includegraphics[width=\linewidth]{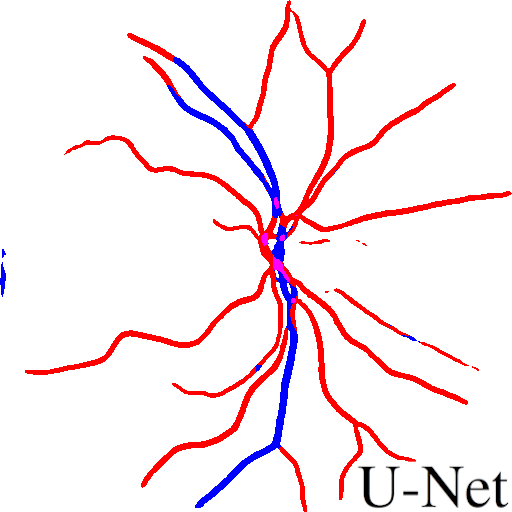}}
\end{minipage}
\label{comparison:b}}\hfil
% Column 3
\subfloat[M0 + diasys + corr.]{
\begin{minipage}[c]{2.7cm}
    \frame{\includegraphics[width=\linewidth]{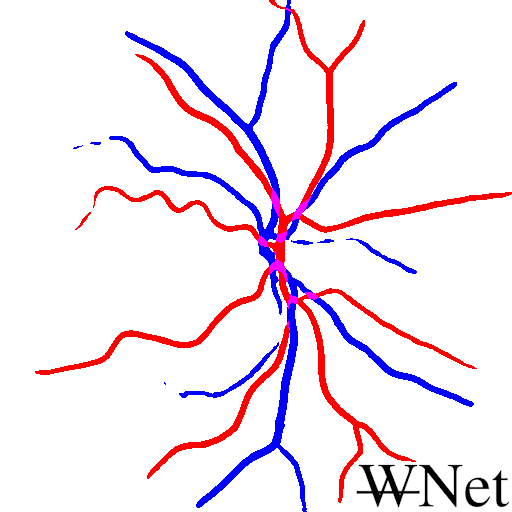}}\\[0.3em]
    \frame{\includegraphics[width=\linewidth]{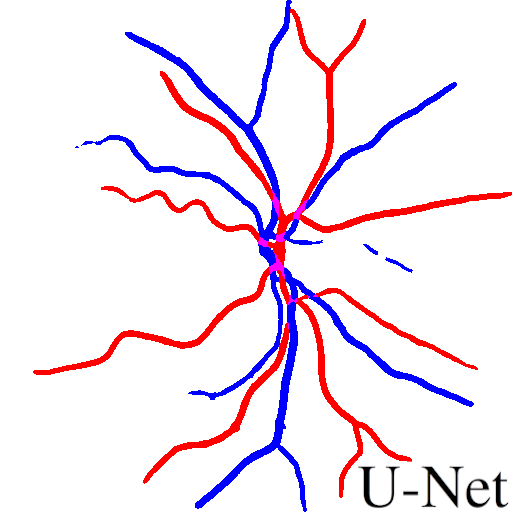}}
\end{minipage}}\hfil

% Column 4
% \subfloat[Boxplot]{
% \begin{minipage}[c]{8.5cm}
%     \frame{\includesvg[width=\linewidth]{figures/unet_av.svg}}
% \end{minipage}
% }
% 
% \subfloat[M0 + correlation]{
% \begin{minipage}[c]{2.8cm}
%     \frame{\includegraphics[width=\linewidth]{figures/pred_mask_corr_white.png}}\\[0.3em]
%     \frame{\includegraphics[width=\linewidth]{figures/pred_unet_corr_white.png}}
% \end{minipage}}\hfil
% % Column 5
% \subfloat[M0 + diasys]{
% \begin{minipage}[c]{2.8cm}
%     \frame{\includegraphics[width=\linewidth]{figures/pred_mask_sys_white.png}}\\[0.3em]
%     \frame{\includegraphics[width=\linewidth]{figures/pred_unet_sys_white.png}}
% \end{minipage}}\hfil
% % Column 6
% \subfloat[Diasys + correlation]{
% \begin{minipage}[c]{2.8cm}
%     \frame{\includegraphics[width=\linewidth]{figures/pred_mask_sys_corr_no_M0_white.png}}\\[0.3em]
%     \frame{\includegraphics[width=\linewidth]{figures/pred_unet_no_m0_white.png}}
% \end{minipage}}

\caption{Segmentation results of \st{W}Net (top) and U-Net (bottom) trained with different inputs. With only M0, the U-Net shows more misclassifications than \st{W}Net (b). Both improve greatly when using the diasys and correlation maps, yielding nearly identical outputs (c).}
\label{comparison}
\end{figure}

\subsection{Direct segmentation}

The most straightforward approach consists in directly performing semantic segmentation on M0 images. The results of this strategy are reported in Table \ref{benchmark}. Overall, all models achieve limited performance, with \st{W}Net obtaining the best results. Most architectures successfully delineate the retinal vasculature but fail to accurately distinguish between arteries and veins (Fig~\ref{comparison:b}), resulting in better segmented arteries, likely due to the higher signal-to-noise ratio and more pronounced pulsatile Doppler signature of arterial flow (Table \ref{ablation}). Other models such as SwinUnet even struggle to segment vessels precisely. Only \st{W}Net manages to perform a consistent classification of arteries and veins (Fig \ref{comparison:b}); this can be attributed to its hybrid design, where convolutional layers ensure accurate spatial segmentation, while transformer layers capture the global context useful for classification. In general, the best-performing models rely on hybrid transformer–convolution designs (\st{W}Net, CSTA-Net) or feature enhancement (MSUNet, UNet++, CE-Net). However, the basic UNet still outperforms several of these more complex models and remains a strong baseline.

\begin{table*}[t]
\footnotesize
\addtolength{\tabcolsep}{-0.15em}
\begin{tabular}{rl|cccc|cccc|r|l}
\hline
\multicolumn{2}{c|}{\multirow{2}{*}{Models}} & \multicolumn{4}{c|}{A/V with M0} & \multicolumn{4}{c|}{A/V with temporal cues} & \multicolumn{1}{c|}{\multirow{2}{*}{\begin{tabular}[c]{@{}c@{}}Params\\ (K)\end{tabular}}} & \multicolumn{1}{c}{\multirow{2}{*}{\begin{tabular}[c]{@{}c@{}}Inf.\\ (ms)\end{tabular}}} \\
\multicolumn{2}{c|}{} & Sens. (\%) & Dice (\%) & clDice (\%) & HD95↓ & Sens. (\%) & Dice (\%) & clDice (\%) & HD95↓ & \multicolumn{1}{c|}{} & \multicolumn{1}{c}{} \\ \hline
UNet & 2015 & 61.5 ± 6.1 & 63.7 ± 4.4 & 67.3 ± 4.9 & 49.80 ± 2.97 & 83.2 ± 1.6 & 82.6 ± 1.0 & 90.8 ± 1.0 & 8.99 ± 1.06 & 17,258 & 16 \\
UNet++ & 2018 & 66.5 ± 7.3 & 67.7 ± 5.8 & 72.0 ± 6.2 & 39.42 ± 0.75 & \textbf{87.4 ± 1.7} & 84.3 ± 0.2 & \textbf{92.7 ± 0.3} & 5.99 ± 1.02 & 9,156 & 22 \\
CE-Net & 2019 & {\ul 68.5 ± 5.3} & {\ul 68.5 ± 6.3} & {\ul 73.5 ± 4.2} & {\ul 36.01 ± 7.72} & {\ul 84.9 ± 2.7} & \textbf{84.5 ± 0.5} & {\ul 92.4 ± 0.7} & 7.17 ± 1.34 & 68,801 & 12 \\
MSUNet & 2021 & 64.5 ± 7.4 & 67.5 ± 6.6 & 71.8 ± 8.4 & 40.64 ± 12.18 & 80.0 ± 0.4 & 81.7 ± 1.3 & 89.6 ± 1.3 & 10.95 ± 3.09 & 47,076 & 28 \\
IterNet & 2020 & 61.0 ± 2.1 & 61.0 ± 1.9 & 63.6 ± 2.2 & 51.72 ± 13.12 & 78.3 ± 4.8 & 79.8 ± 2.5 & 87.7 ± 2.6 & 16.53 ± 6.05 & 17,476 & 4 \\
WNet & 2025 & 48.9 ± 1.6 & 46.3 ± 1.5 & 49.1 ± 1.2 & 82.31 ± 3.84 & 79.7 ± 3.0 & 81.1 ± 1.0 & 88.6 ± 0.8 & 10.17 ± 1.15 & 69 & 6 \\
RRWNet & 2024 & 46.6 ± 7.2 & 50.2 ± 8.7 & 53.8 ± 7.6 & 71.26 ± 24.7 & 81.9 ± 4.6 & 81.9 ± 1.0 & 89.9 ± 1.3 & 9.54 ± 2.96 & 62,063 & 132 \\
SwinUNet & 2022 & 40.6 ± 10.5 & 47.5 ± 8.9 & 49.0 ± 12.7 & 64.22 ± 15.2 & 66.0 ± 5.3 & 72.2 ± 3.3 & 77.5 ± 3.7 & 23.33 ± 4.31 & 43,077  & 30 \\
\st{W}Net & 2025 & \textbf{72.9 ± 1.0} & \textbf{73.3 ± 1.9} & \textbf{78.7 ± 2.1} & \textbf{30.35 ± 5.33} & 84.2 ± 2.0 & 84.0 ± 0.5 & 92.3 ± 0.5 & {\ul 5.98 ± 0.44} & 7,034 & 18 \\
CSTA-Net & 2025 & 59.4 ± 1.0 & 61.0 ± 0.9 & 63.4 ± 1.1 & 42.69 ± 0.5 & 83.1 ± 0.2 & {\ul 84.4 ± 0.3} & 91.9 ± 0.3 & \textbf{5.61 ± 0.29} & 21,947 & 31 \\ \hline
\end{tabular}
\caption{Segmentation performance of traditional and state-of-the-art methods on two tasks: direct artery/vein (A/V) segmentation from M0 and A/V segmentation using temporal cues from the sequential pipeline (Fig. \ref{fig:segmentation}). Results are reported as mean ± standard deviation over five training runs. Sensitivity, Dice (F1), and clDice \cite{clDice} are expressed in percentage, and the 95th-percentile Hausdorff distance in arbitrary units. All scores are averaged over artery and vein classes. Best and second-best results are shown in bold and underlined, respectively.}
\label{benchmark}
\end{table*}

\begin{table}[]
\footnotesize
\addtolength{\tabcolsep}{-0.1em}
\begin{tabular}{@{}llllll@{}}
\toprule
\multicolumn{1}{c}{} & \multicolumn{1}{c}{M0} & \multicolumn{1}{c}{\begin{tabular}[c]{@{}c@{}}M0 + corr.\\ + diasys\end{tabular}} & \multicolumn{1}{c}{M0 + corr.} & \multicolumn{1}{c}{\begin{tabular}[c]{@{}c@{}}M0 \\ + diasys\end{tabular}} & \multicolumn{1}{c}{\begin{tabular}[c]{@{}c@{}}Diasys \\ + corr.\end{tabular}} \\ \midrule
Artery & 65.0 ± 8.2 & \textbf{83.4 ± 5.9} & 82.1 ± 4.9 & 75.0 ± 5.9 & 72.7 ± 7.6 \\
Vein & 55.0 ± 9.9 & \textbf{82.4 ± 5.5} & 80.8 ± 6.7 & 75.7 ± 10.9 & 45.6 ± 18.3 \\ \bottomrule
\end{tabular}
\caption{Ablation study of the segmentation pipeline. Average Dice score and standard deviation for artery and vein segmentation with a U-Net trained with M0 alone or with aditionnal temporal cues. Temporal information improves accuracy, reduces variability—particularly for veins—and narrows the artery–vein performance gap. The correlation map provides most of the discriminative power, but optimal results are achieved when combined with diasys, while M0 remains essential for spatial and topological information.}
\label{ablation}
\end{table}

\subsection{Sequential segmentation strategy based on cardiac pulse signal analysis} \label{sequential}

The first approach performs poorly because it relies solely on spatial information and disregards the temporal dynamics inherent to our data. A more effective strategy exploits the cardiac pulse signal present in the Doppler Power sequence to distinguish veins from arteries. The overall pipeline, illustrated in figure \ref{fig:segmentation}, is summarized as follows:

\begin{itemize}
\item A binary blood vessel segmentation is first performed using Power Doppler images as input, with a standard UNet achieving a 0.82 Dice score.

\item The vessel masks are then skeletonized, and junction points are removed to obtain distinct vessel segments. For each labeled segment, the mean temporal signal is computed across all frames. Vessels exhibiting peaks with a positive derivative exceeding a predefined threshold are classified as arteries. Although this procedure results in a number of false negatives, it effectively identifies arteries displaying the most pronounced systolic response.

\item This preliminary mask is used to estimate the global cardiac pulse signal across all frames. The zero-lag correlation between each pixel's signal and the global average signal is then computed for each frame. Average systolic and diastolic frames are obtained by averaging all frames around the maximum and minimum peaks, respectively.

\item Finally, artery–vein segmentation is performed by providing the Power Doppler image, the correlation map, and the diasys image (defined as the systolic frame minus the diastolic frame) as inputs to the model. The diasys image and the correlation map convey similar information, but they differ and can be complementary, as shown in fig~\ref{inputs}. Table~\ref{ablation} shows their impact on training.
\end{itemize}

The results are summarized in Table~\ref{benchmark}. The proposed strategy substantially improves performance across all models, reduces variability, particularly for vein segmentation, and eliminate the performance gap observed between arteries and veins when using M0 alone (Table ~\ref{ablation}). Previously, only the \st{W}Net achieved a Dice score above 0.7, exceeding the second-best model by more than 0.05 and the standard UNet by about 0.15. With the proposed approach, almost all models now reach Dice scores above 0.8, within 0.03 of each other, with similar trends across other metrics.  This indicates that architectures originally designed for spatial feature extraction can effectively exploit temporal information encoded in the correlation and diasys maps (Fig.~\ref{comparison}).

Notably, while the basic UNet was previously outperformed by more advanced architectures when trained solely on static M0 inputs, it now achieves results comparable to the best models once temporal cues included. The same goes for the UNet++, which has the best sensitivity and best clDice score, the CE-Net, which now achieves best or second best score on all metrics, or the WNet, which achieves competitive performance with much less parameters. It seems that by using discriminative temporal information, performance become more uniform across models, reducing the relative benefit of complex designs. SwinUnet is the only model that moderately benefits from the proposed temporal cues and remains significantly below other architectures, suggesting that convolutional inductive biases are still essential for accurate retinal vessel segmentation.

\section{Conclusion}

Doppler holography uniquely combines wide-field imaging of retinal vasculature with quantitative, time-resolved measurements of blood flow. This capability opens new possibilities for assessing micro-vascular health by linking vessel morphology with hemodynamic behavior. However, extracting meaningful flow information requires accurate artery–vein segmentation, a task complicated by low contrast, overlapping choroidal vessels, and variable flow signals. Traditional, spatially focused segmentation strategies—effective in fundus or OCT images—prove less suitable for Doppler holography, where temporal information carries the most diagnostic value. By leveraging the cardiac pulse dynamics embedded in the Doppler signal, arteries and veins can be distinguished more robustly. Using this approach, we demonstrate that even simple U-Net-like architectures, when enhanced with temporal features, can rival the performance of far more complex models, underscoring the critical role of time-resolved analysis in Doppler holographic imaging. These findings emphasize the need for new deep learning frameworks explicitly tailored to the temporal dynamics of Doppler holographic signals, paving the way for more effective and interpretable processing of time-resolved holographic data.

\bibliographystyle{IEEEbib}
% \bibliography{refs,segmentation_models}
\bibliography{biblio}

\end{document}